\documentclass[runningheads]{llncs}
\usepackage{graphicx}
\usepackage{subfig}
\usepackage[round]{natbib}
\usepackage{hhline}
\usepackage{booktabs} 
\newcommand{\ra}[1]{\renewcommand{\arraystretch}{#1}}
\usepackage{siunitx}

\begin{document}

\title{Multi-View Semantic Labeling of 3D Point Clouds for Automated Plant Phenotyping}

\titlerunning{Semantic Labeling of 3D Point Clouds for Automated Plant Phenotyping}

\author{Bernhard Japes \inst{1} \and
Jennifer Mack \inst{1} \and
Florian Rist \inst{2} \and
Katja Herzog \inst{2} \and
Reinhard T\"opfer \inst{2} \and
Volker Steinhage \inst{1} }

\authorrunning{B. Japes et al.}

\institute{Department of Computer Science IV, University of Bonn, Endenicher Allee 19a, D-53115 Bonn, Germany \and
Institute for Grapevine Breeding Geilweilerhof, Julius K\"uhn-Institut,
Federal Research Centre for Cultivated Plants, Siebeldingen, Germany}

\maketitle

\begin{abstract}
Semantic labeling of 3D point clouds is important for the derivation of 3D models from real world scenarios in several economic fields such as building industry, facility management, town planning or heritage conservation. In contrast to these most common applications, we describe in this study the semantic labeling of 3D point clouds derived from plant organs by high-precision scanning. Our approach is optimized for the task of plant phenotyping with its very specific challenges and is employing a deep learning framework. Thereby, we report important experiences concerning detailed parameter initialization and optimization techniques. By evaluating our approach with challenging datasets we achieve state-of-the-art results without difficult and time consuming feature engineering as being necessary in traditional approaches to semantic labeling.

\end{abstract}

\section{Introduction}
\subsection{Motivation}

In plant breeding, phenotyping refers to the measuring and evaluation of observable plant traits over time, aiming at the determination of yield potential, stress resistance and crop quality. Traditionally, phenotyping is done manually. Therefore, phenotyping is highly labor-intensive and costly, resulting in the so-called "phenotyping bottleneck" \citep{FurbankTester}. Current research addresses this bottleneck by employing innovative technologies for the automatic derivation of plant traits.

Within an interdisciplinary research project of the Computer Science Department of Bonn University and the Julius K\"uhn Institute for Grapevine Breeding we investigate and develop fully automated approaches for sensor-based high-throughput derivation of objective and high-quality phenotypic data. Within this project, we focus on sensors that deliver 3D data in terms of so-called point clouds. 3D data allows for automated reconstruction of complete 3D models of given plants or plant organs, providing the opportunity to derive arbitrary phenotype traits in a direct way in contrast to approximate estimations from pure 2D image data.

A crucial step in sensor data interpretation is the semantic labeling, i.e., the process of classifying all sensor data points into semantic categories. In this application semantic categories are plant organs like twigs, pedicels, berries, etc. The semantic labeling of point cloud data is a complex problem especially in the case of grape bunch segmentation. While some approaches exist that differentiate parts of berries and parts of the stem \citep{Paulus2013, Mack17} the semantic labeling of 3D point clouds taken from the stem skeleton of a grape bunch has, to the best of our knowledge, not been tried so far. This semantic labeling will be used as input for a more detailed reconstruction of the stem skeleton from which traits like lengths and diameter of the different parts of the skeleton could be derived.

\subsection{Contributions and overview}
\label{sec:CaO}

\begin{figure}[!htbp]
  \centering
  \subfloat[RGB point cloud]{\includegraphics[width=0.2\textwidth]{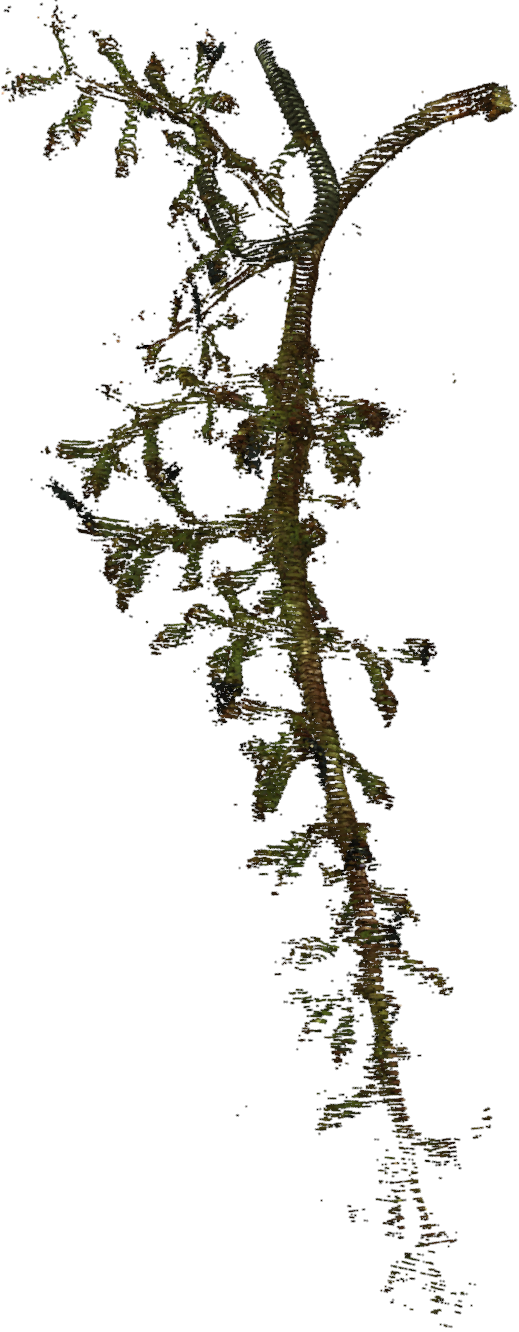}\label{fig_n5a}}
  \hfill
  \subfloat[Labeling scheme]{\includegraphics[width=0.46\textwidth]{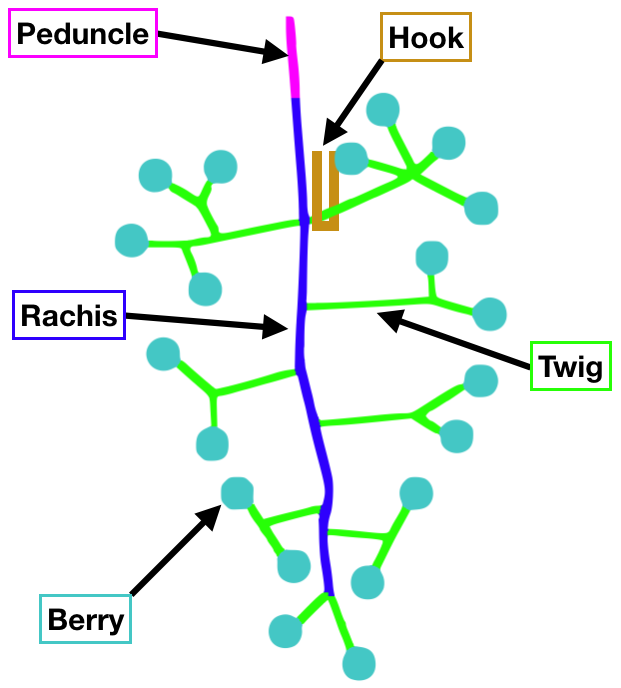}\label{fig_n5b}}
  \hfill
  \subfloat[Ground truth]{\includegraphics[width=0.2\textwidth]{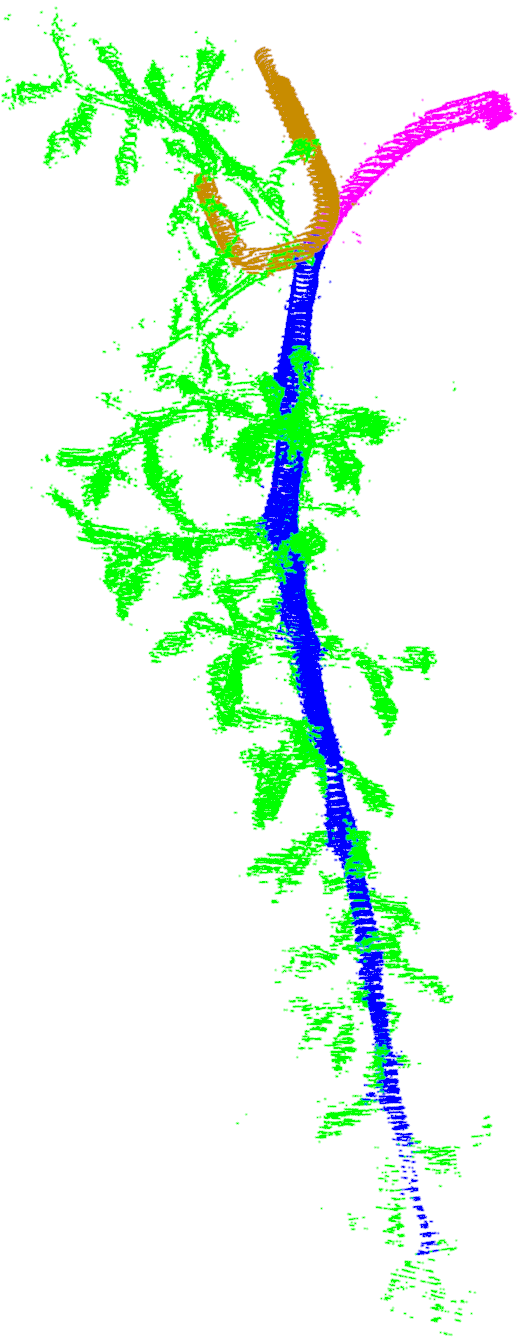}\label{fig_n5c}}
  \caption{The classes used in our semantic labeling of grape bunches. We take our RGB point clouds of grape bunches a) and apply the labeling scheme b) to create the labeled ground truth c).}
  \label{fig_nom5}
\end{figure}

Instead of using a traditional semantic labeling approach, we employ a deep learning approach. While traditional semantic labeling approaches require serious feature engineering, i.e., the employment of hand-crafted feature descriptors to describe the sensor data, deep learning approaches learn these feature descriptions from training data. In this study, we employ an adjusted version of the SnapNet approach \citep{SnapNet} for the application of grape bunch segmentation. We distinguish between the following classes (see also Figure \ref{fig_nom5}): the peduncle (the former connection of the grape bunch to the branch of the grapevine plant), the rachis (the main stem of the bunch), the berries, twigs (including larger and smaller twigs) and the hook (used to hang the bunch for scanning). This study shows three contributions. First, we present the adjustments of the SnapNet approach for this phenotyping application. Second, we present an evaluation of the derived semantic labeling results. Third, we discuss the results as well as additional options and future work.

\section{Related Work}
\label{sec:RW}

Generally, semantic labeling is a popular and up-to-date research field. The 2D task of giving a pixel-wise labeling of images has already been studied intensively since the early 2000s within challenges like the Pascal Visual Object Classes Challenge \citep{PascalVOC}. Currently the CNN-based encoder-decoder architectures like SegNet \citep{SegNet} and U-Net \citep{UNet2D} provide efficient solutions.

However, the 3D task of giving a point-wise labeling of so-called 3D point clouds has only recently gained interest with the increasing availability of laser scanners and possible applications in the field of robotics.
In order to compute such a point-wise classification the traditional approaches use handcrafted features derived from the points and their local neighborhood \citep{Weinmann} with properties like fast point feature histograms \citep{FPFH}. However, over the last couple of years deep learning approaches to semantic labeling of 3D point clouds have been proposed.
One of the most important benchmarks in order to compare these new approaches has been the Semantic 3D benchmark \citep{Sem3D} with over four billion hand-labelled 3D-points of urban outdoor scenes. Several approaches have been optimized for this specific dataset and yet, to the best of our knowledge, there is no deep learning approach to the semantic labeling in the field of phenotyping plant traits.

Currently the most promising deep learning approaches for urban outdoor scenes can be grouped into three categories: voxel-based methods, point-based methods, and multi-view methods.

\subsection{Voxel-Based Methods}
The voxel-based methods use voxel grids to generate a regular neighborhood structure. This allows CNNs to operate on voxels similar to the way pixels are processed during image classification \citep{VoxN}. For small point clouds it is possible to use end-to-end 3D CNN approaches like 3D U-Net \citep{UNet3D}. In this case the existing 2D architecture of U-Net \citep{UNet2D} has been adapted to 3D input and output data by using multiple layers of 3D convolutions, 3D max pooling and 3D up-convolutions. However, due to the large amount of parameters end-to-end 3D CNN approaches are not feasible for larger point clouds.
\citet{Sem3D} proposed a 3D CNN Baseline that is using a sliding 3D window approach based on the ideas of \citet{VoxN} and \citet{ShapeN}.
For each point they compute voxel occupancy grids of different scales as inputs to multiple CNNs. By doing so they are able to capture the same degree of details regardless of the size of the point cloud. Nevertheless this approach requires very powerful GPUs with a lot of memory to work. For our task of plant phenotyping we are looking for an approach that can also be used on less powerful hardware.

\subsection{Multi-View Approaches}
The multi-view approaches operate on 2D images of the given 3D point cloud. \citet{su15} first presented the Multi-View CNN which is able to classify objects based on a set of rendered images and multiple CNNs. \citet{SnapNet} went a step further and introduced SnapNet, a semantic labeling based on deep segmentation networks. In this case sets of 2D images of the scene are generated and processed through U-Net to create labeled 2D images. In a postprocessing step those labeled views are projected back onto the original point cloud to compute a dense 3D point labeling. They further refined their approach and proposed SnapNet-R \citep{SnapNetR} which is optimized for robotics, where RGB-D images are common.

\subsection{Point-Based Methods}
The point-based methods operate directly on point clouds without the need of voxels or 2D representations. Architectures like PointNet use fully-connected classification and segmentation networks to compute a semantic labeling \citep{PointN}. The further refined PointNet++ introduced a hierarchical neural network and recursive applications of PointNet to capture local structures at different scales \citep{PointNpp}.
\citet{SuperP} propose a method that combines a partitioning into geometrically simple shapes with PointNet embeddings.

\section{Methods and Data}
\label{sec:MaD}

\subsection{Data}
\label{sec:Data}

Our data has been captured with the Artec Space Spider, a structured light 3D scanner with a resolution of 0.1mm and an accuracy of 0.05mm.
We generated point clouds for four different cultivars: Calardis blanc, Dornfelder, Pinot noir and Riesling. The number of acquisitions and the average size of point clouds per cultivar are shown in table \ref{tab:AvgSize}. First we scanned each grape bunch with its berries. To capture the stem structure in full detail we later removed the berries and scanned the grape bunch once more.
The grape bunches of cultivar Riesling have many small virgin berries, which can be removed only very laboriously. For that reason the stem structure scans of the cultivar Riesling would have been quite confusing and it would be very hard to define a reliable ground truth.
In total we have created a dataset consisting of 33 point clouds that have been manually labeled according to the labeling scheme presented in figure \ref{fig_nom5}.

\begin{table*}[!t]\centering
\caption{Number of acquisitions and the average size of point clouds per cultivar}
\label{tab:AvgSize}
\ra{1.3}
\begin{tabular}{@{}lccrccrccrccr@{}}\toprule
& \phantom{ab} & \multicolumn{2}{c}{Calardis blanc} & \phantom{ab}& \multicolumn{2}{c}{Dornfelder} &
\phantom{ab} & \multicolumn{2}{c}{Pinot noir} & \phantom{ab} & \multicolumn{2}{c}{Riesling} \\ \cmidrule{3-4} \cmidrule{6-7} \cmidrule{9-10} \cmidrule{12-13}
& & \# & size  && \# & size  && \# & size && \# & size \\ \midrule
stem structure only & & 4 & 239215  && 5 & 218420 && 5 & 148355 && 0 & N/A\\
grape bunch with berries & & 4 & 1040599  && 5 & 994116 && 5 & 677241 && 5 & \phantom{} 183328\\ \bottomrule
\end{tabular}
\end{table*}

\subsection{Workflow}
\label{sec:Workflow}

Since the current voxel-based methods are too computationally expensive for our high-resolution data and the current point-based methods are struggling with complex scenes at different scales, which are very common in plant phenotyping, we have decided to base our model on state-of-the-art multi-view approaches.

\begin{figure}[!h]
\includegraphics[width=\textwidth]{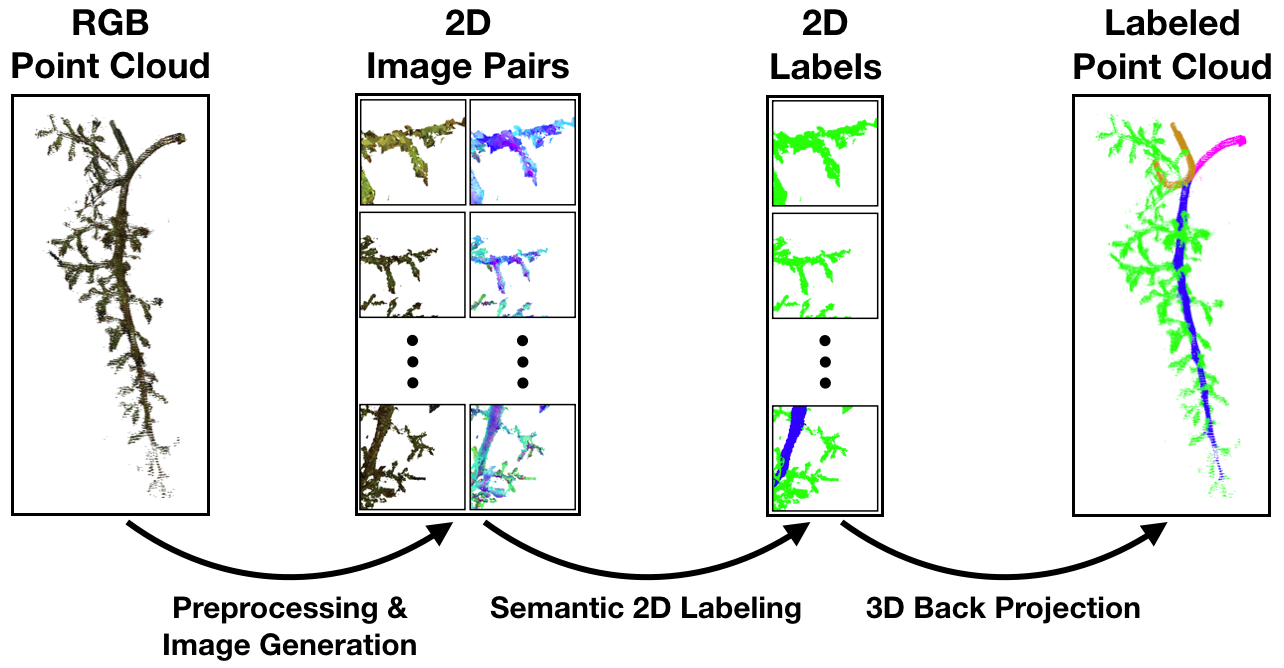}
\caption{The overall workflow of our multi-view semantic labeling. We perform a preprocessing of the 3D point cloud and generate 2D image pairs. Based on these 2D images we compute multiple semantic 2D labels. Finally we perform a 3D back projection of these 2D labels to create a labeled 3D point cloud.}
\label{fig_graphabstract}
\end{figure}

By utilizing multiple 2D images with great variety in camera positions and angles we are able to construct a sufficient representation of our 3D point clouds. These 2D images can be efficiently processed by CNNs with minimal GPU memory requirements, which facilitates practical use. 
The composed workflow is shown in figure \ref{fig_graphabstract}. Like other multi-view based approaches our model relies on three fundamental steps:

\begin{enumerate}
\item{We perform a preprocessing of the RGB point cloud and \textbf{generate the 2D images}.}
\item{A \textbf{semantic labeling} of the 2D images is computed and the 2D scores are stored.}
\item{We use the stored 2D scores and perform a \textbf{3D back projection} to create the labeled point cloud.}
\end{enumerate}

The first step in our processing pipeline is the preprocessing. To reduce the point cloud size and create a homogeneous point cloud we voxelize the space and keep the closest point to each voxel center. This simplifies the further processing steps and reduces the runtime significantly. Based on this voxelized point cloud we can generate a RGB-mesh as a dense representation of the object that can be used to generate meaningful images.
In order to maintain some kind of volume information during our image based semantic labeling we generate an additional composite-mesh that will provide additional input images. Like \citet{SnapNet} we encode the normal deviation on the green channel and the local noise on the red one. 

Based on these two meshes we can generate image pairs for the 2D semantic labeling. To cover as much 3D information as possible in our images we have to choose our camera positions wisely. \citet{SnapNet} proposed a multiscale strategy to solve this problem. They randomly pick a point of the scene, create a line that goes through the point and generate three camera positions on this line facing towards the point itself. This procedure is applied to both meshes. 

An RGB image is generated based on the color information of our RGB mesh. We further generate a composite image by using the precomputed normal deviation on the green channel, the precomputed local noise on the red channel and the distance to the camera on the blue channel. Those two images combined as an image pair can be used as an input to our 2D semantic labeling.

\begin{figure}[!h]
\includegraphics[width=\textwidth]{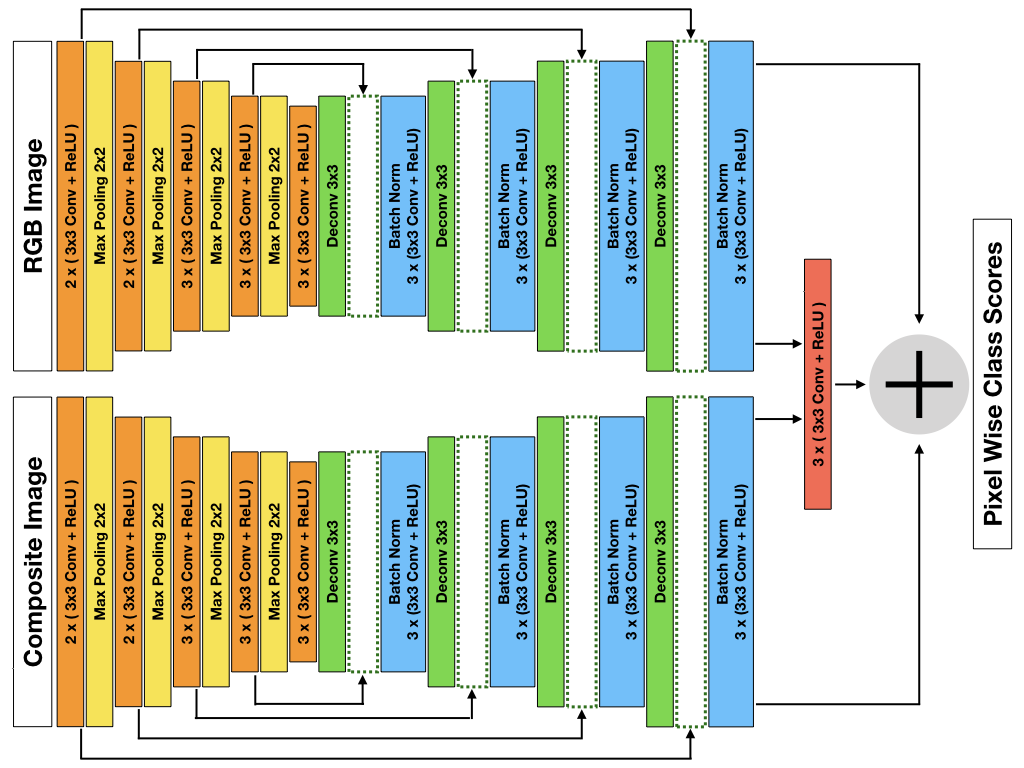}
\caption{The U-Net based fusion network with residual correction used in this paper from \citet{SnapNet}. The arrows indicate concatenations of layers.}
\label{fig_unet}
\end{figure}

As mentioned in section \ref{sec:RW} the semantic labeling of 2D images has been studied intensively. Several CNN-based encoder-decoder architectures like U-Net \citep{UNet2D} already provide efficient solutions. The architecture of our model is shown in Fig. \ref{fig_unet}:
We use an U-Net-based fusion network with residual correction from \citet{SnapNet} to combine RGB and composite information. For a detailed layer-wise explanation of the architecture the reader may refer to the freely available articles by \citet{UNet2D} and \citet{SnapNet}.

The architecture can be trained in three individual steps based on the generated RGB, composite and ground truth images. First we independently train the U-Net parts for RGB and composite images based on a pretrained VGG-16 encoder \citep{vgg16}. Then these trained U-Nets are used to compute the inputs for our residual correction and train the fusion network.

During testing the output of our model is supposed to be a labeled point cloud. This can be achieved by processing multiple images and projecting the pixel wise class scores back on to the mesh that has been used to create the image. The score vectors are added and the final label of a vertex in our mesh is the respective class label with the highest score. Based on these labeled vertices we assign a class label to each point of our original point cloud by choosing the class label of the nearest neighbor.

\subsection{Optimizations/Adaptions/Implementation Details}
\label{sec:OAID}

The research of semantic point cloud labeling focused mainly on large point clouds of outdoor scenes. Characteristic for these point clouds is: a) a very high number of points, b) multiple flat surfaces created by the ground or buildings and c) a similar structure with the ground (man made or natural) at the bottom of the scene.

Our model is specifically optimized for the task of plant phenotyping with its very own challenges. The point clouds that we are operating on are typically high-resolution point clouds of relatively small objects with complex shapes. To cover all areas of our objects we generate a lot more images with greater variation in position. A detailed comparison of the popular Semantic 3D dataset \citep{Sem3D} used by \citet{SnapNet} and our custom dataset is shown in table \ref{tab:Data}.

\begin{table*}[!t]
\centering
\caption{Comparison of the urban scenes of the Semantic 3D dataset \citep{Sem3D} as used by SnapNet \citep{SnapNet} and our plants}
\label{tab:Data}
\ra{1.3}
\begin{tabular}{@{}lcrcr@{}}\toprule
& & \multicolumn{1}{c}{Urban} & & \multicolumn{1}{c}{Plants} \\ \midrule
\underline{Number of Point Clouds} \\
Total && 15 && 33 \\
Training                          && 9         && 19      \\
Test                               && 6         && 14      \\ \midrule
\underline{Size of Point Clouds} \\
Min                               && 16658648  && 75975   \\
Average                           && 110686366 && 491714  \\
Max                               && 280994028 && 1552774 \\ \midrule
\underline{Number of Image Pairs at Training} \\ 
Per Point Cloud                   && 400       && 1800    \\
Total                             && 3600      && 34200   \\ \midrule
\underline{Number of Image Pairs at Testing}        \\
Per Point Cloud                   && 1500      && 1500    \\
Total                             && 9000      && 21000   \\ \bottomrule
\end{tabular}
\end{table*}

Similar to \citet{SnapNet} we use a multiscale strategy with adjusted distances. The final parameters of our model are listed in table \ref{tab:Param}. Our experiments have shown that the choice of distances is of great importance in plant phenotyping. If the camera is too close to a grape bunch the distinction between twigs, rachis and peduncle is nearly impossible due to the lack of context information and the high degree of self-similarity. We should further try to avoid camera positions that are too far away from our grape bunches since the outer twigs tend to cover most of the inner stem structure.
Instead we choose the distance in relation to the size of a grape bunch as indicated by the radius of a grape bunch. For the grape bunches in our dataset with a radius between \SI{63.9}{mm} and \SI{130.4}{mm} we found that images with camera distances of \SI{20}{mm}, \SI{40}{mm} and \SI{60}{mm} provide insightful input data to our network.

\begin{table*}[!h]
\centering
\caption{Properties of our data and respective parameter choices}
\label{tab:Param}
\ra{1.3}
\begin{tabular}{@{}lcl@{}}\toprule
Scan Resolution             && \SI{0.1}{mm}                          \\
Voxel Size                  && \SI{0.4}{mm}                        \\ \midrule
\multicolumn{3}{l}{Radius of Grape Bunches}                                \\
Min                         && \SI{63.9}{mm}                       \\
Average                         && \SI{94.5}{mm}                       \\
Max                         && \SI{130.4}{mm}                      \\ \midrule
View Distances              && \SI{20}{mm}, \SI{40}{mm}, \SI{60}{mm}           \\
View Azimuth                && chosen uniformly at random $\mathcal{U}$(0$^{\circ}$,360$^{\circ}$)  \\
View Elevation              && chosen uniformly at random $\mathcal{U}$(-90$^{\circ}$,90$^{\circ}$)  \\ \bottomrule
\end{tabular}
\end{table*}

Whereas near vertical camera angles with an elevation between 70$^{\circ}$ and 90$^{\circ}$
are preferred for outdoor scenes we vary the elevation from -90$^{\circ}$ to 90$^{\circ}$.
This increases the probability that even the mostly concealed areas are captured at least once during training or rather testing and provides some robustness against rotations of stem structures.
Due to the relatively small size of grape bunches the resulting point clouds are also significantly smaller than the point clouds of outdoor scenes used in the Semantic 3D dataset despite the high-resolution. In fact these point clouds of outdoor scenes are more than 200 times larger on average with 110M points compared to the 0.5M points of a grape bunch scan.

By applying a sparse point cloud decimation with a voxel size of \SI{0.1}{m} \cite{SnapNet} reduced the size of the urban scenes down to less than 2.3M points. Due to the large percentage of flat surfaces in the scenes this is possible without loosing too much information. Our complex stem structures can not be sparsified to this extent. Instead we only apply a rather fine voxel size of \SI{0.4}{mm} to create homogenous point clouds with 0.16M points on average.
The significantly smaller number of points in our point clouds simplifies the image generation and enables the generation of far more images for training in a reasonable time. Instead of 400 image pairs per point cloud as used on the Semantic 3D dataset we generate 1800 image pairs per point cloud, which means that our network is trained on 34200 image pairs in total.

Combined with the large variety of camera positions this has shown to provide a decent amount of generalization, even on a small set of training data. In practice this is very important as the manual labeling of training samples is known to be labor-intensive and costly, especially on such complex shapes.

\section{Experiments and results}

\subsection{Evaluation}

We evaluate the model on our custom dataset as described in section \ref{sec:Data} with five different classes: rachis, berries, peduncle, hook and twigs.
Our dataset consists of 33 different grape bunches of the cultivars Calardis blanc, Dornfelder, Pinot noir and Riesling (19 for training and 14 for testing). The experiments were evaluated on a system with Intel Core i5 6600k 4x 3.50GHz, 16GB DDR4 RAM and GeForce GTX 1070.

For training we performed the preprocessing as described in section \ref{sec:OAID} and generated 1800 image pairs per point cloud thus 34200 image pairs in total. The neural networks have been trained with a stochastic gradient descent with a learning rate of 0.0001 and a momentum of 0.9 with a batch size of 8 for 10 epochs each. The total runtime for training has been 30h on the system above.

At testing the pixel-wise class scores of multiple images are combined and projected back on to the original point cloud. Due to the random image generation the results get more consistent by using more images. However, the generation and semantic labeling of each individual image pair is rather complex and requires a certain amount of time. Table \ref{tab:IMG} lists the test results and runtimes depending on the number of image pairs for the whole test dataset. The backprojection requires a constant 122 seconds independent of the number of image pairs and is included in the total runtimes.

\begin{table*}[!t]
\centering
\caption{Results for different numbers of test image pairs}
\label{tab:IMG}
\ra{1.3}
\begin{tabular}{@{}lcrcrcrcrcrcrcrc@{}}\toprule

Number of Image Pairs   && \multicolumn{1}{c}{30} && \multicolumn{1}{c}{150} && \multicolumn{1}{c}{300} && \multicolumn{1}{c}{600} && \multicolumn{1}{c}{900} && \multicolumn{1}{c}{1200} && \multicolumn{1}{c}{1500}                          \\ \midrule
\underline{Performance} \\
AIoU  &&  0.737  &&  0.842  &&  0.866  &&  0.866   &&  0.876    &&   0.877  &&  0.877   \\
Accuracy   &&  0.964  &&  0.981   &&   0.984  &&  0.984   &&  0.985   &&   0.985   &&   0.985   \\ \midrule
\underline{Runtime in seconds} \\
Preprocessing \& Image Gen.   &&  178  &&  223   &&  292   &&  449   &&  607   &&   761   &&   923   \\
Semantic Labeling   &&  76  &&  188   &&  318   &&  579   &&  838   &&  1102    &&   1361   \\
Total Runtime &&  376  &&  533   &&  732   &&  1150   &&  1567   &&  1985    &&   2406   \\ \bottomrule
\end{tabular}
\end{table*}

\begin{figure}[!b]
  \centering
  \subfloat[Normalized values]{\includegraphics[height=5cm]{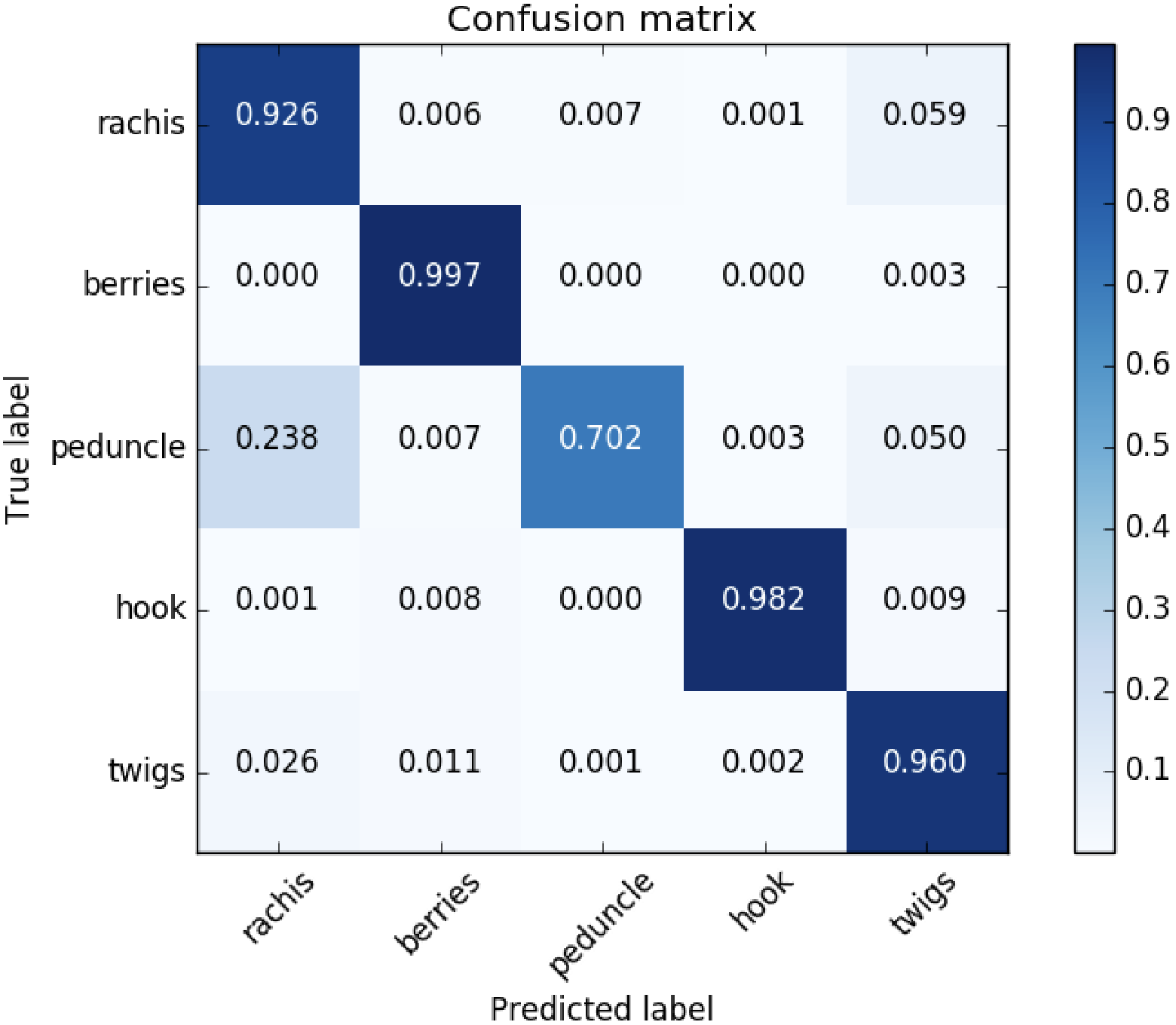}\label{fig_cm5norm}}
  \hfill
  \subfloat[Absolute values]{\includegraphics[height=5cm]{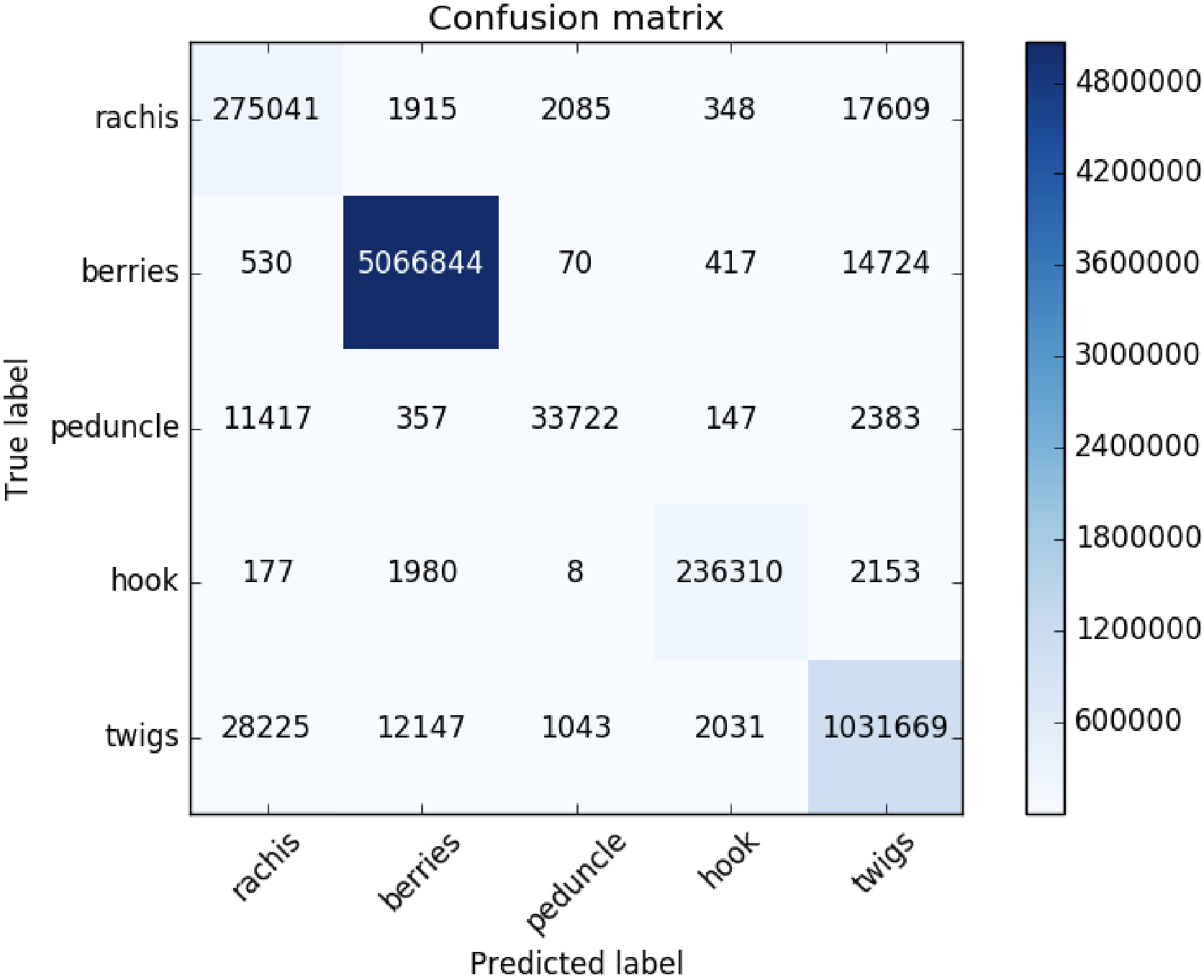}\label{fig_cm5abs}}
  \caption{Confusion matrix of the generated output for the whole set of test data.}
  \label{fig_cm5both}
\end{figure}

It can be seen that AIoU as well as accuracy improve with a higher number of image pairs. Yet the performance does not change significantly for more than 900 image pairs. The runtime however is increasing linearly with the number of image pairs. Because of that we recommend using 900 image pairs in practice, especially if either time or computing power is limited. For our further discussion we will continue with the results for 1500 image pairs to maximize the consistency.

To get a better understanding of the prediction performance it is worth taking a look at the confusion matrix of our model as shown in figure \ref{fig_cm5both}.
It can be seen that our model is able to detect and label rachis, berries, hook and twigs with high accuracies ranging from 0.926 to 0.997. The only significant mistake is the misclassification of points with class peduncle as rachis. 

\begin{figure}[!t]
\includegraphics[width=\textwidth]{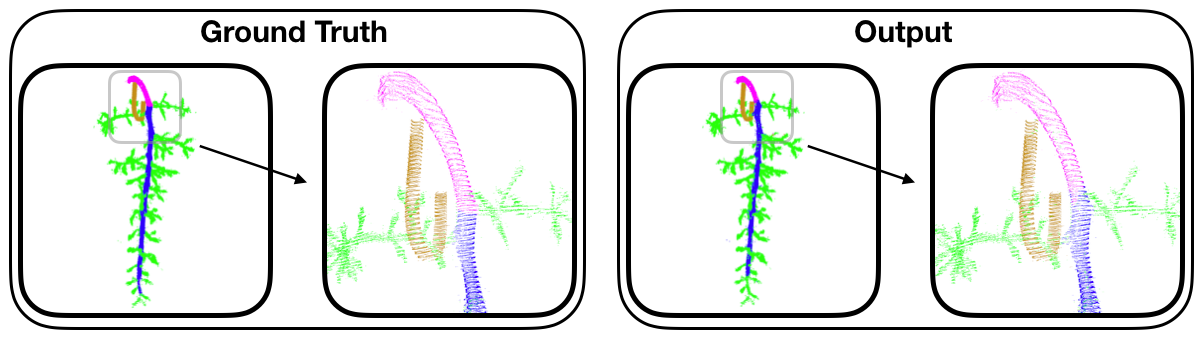}
\caption{Comparison of ground truth and output on a bunch of cultivar Dornfelder without berries. The labels shown are explained in Fig. \ref{fig_nom5}.}
\label{fig_compP}
\end{figure}

\begin{figure}[!b]
\includegraphics[width=\textwidth]{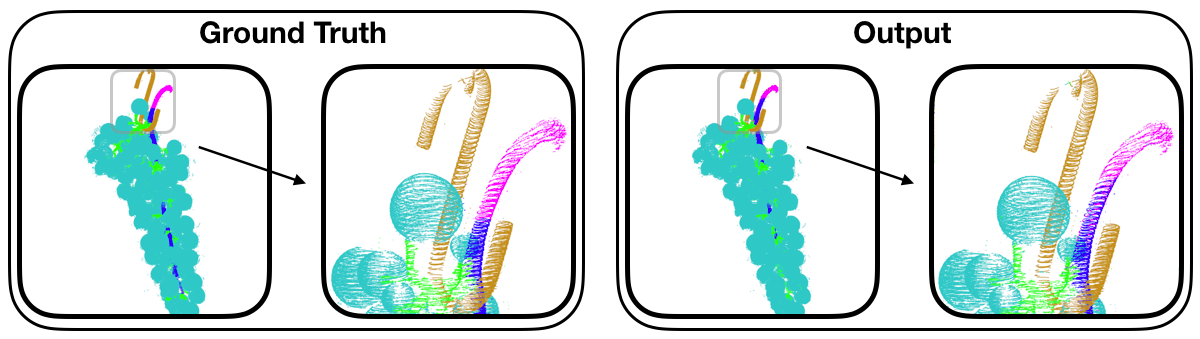}
\caption{Comparison of ground truth and output on a bunch of cultivar Dornfelder with berries. The labels shown are explained in Fig. \ref{fig_nom5}.}
\label{fig_compB}
\end{figure}

\subsection{Discussion}

Figure \ref{fig_compP} compares the ground truth with the output of our model on a grape bunch without berries. We can see that the misclassification of the peduncle is mainly influenced by a different height at which the rachis turns into the peduncle. In practice however it is primarily important that the upper end is correctly identified as peduncle, which our model satisfies. Other than that the predictions are mostly correct apart from some single outliers that are only visible on closer examination.

\begin{figure}[!tbp]
  \centering
  \subfloat[RGB point cloud]{\includegraphics[width=0.2\textwidth]{images/d1.png}\label{fig_n6a}}
  \hfill
  \subfloat[Labeling scheme]{\includegraphics[width=0.46\textwidth]{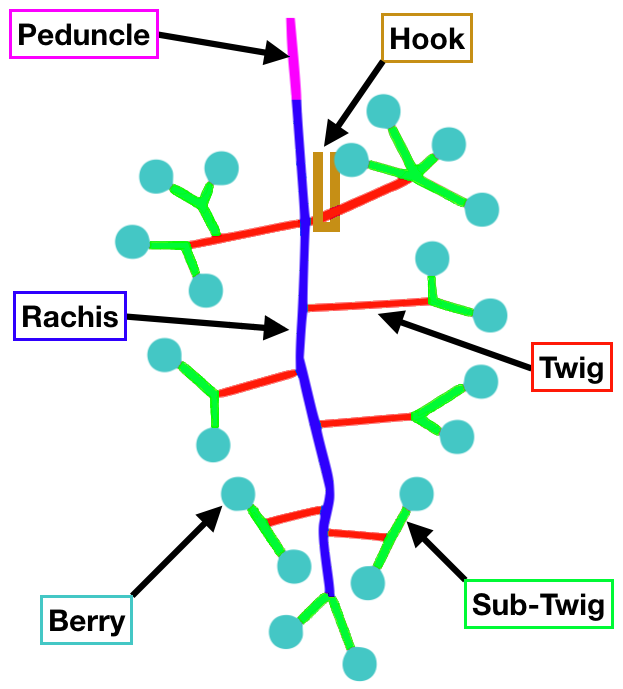}\label{fig_n6b}}
  \hfill
  \subfloat[Ground truth]{\includegraphics[width=0.2\textwidth]{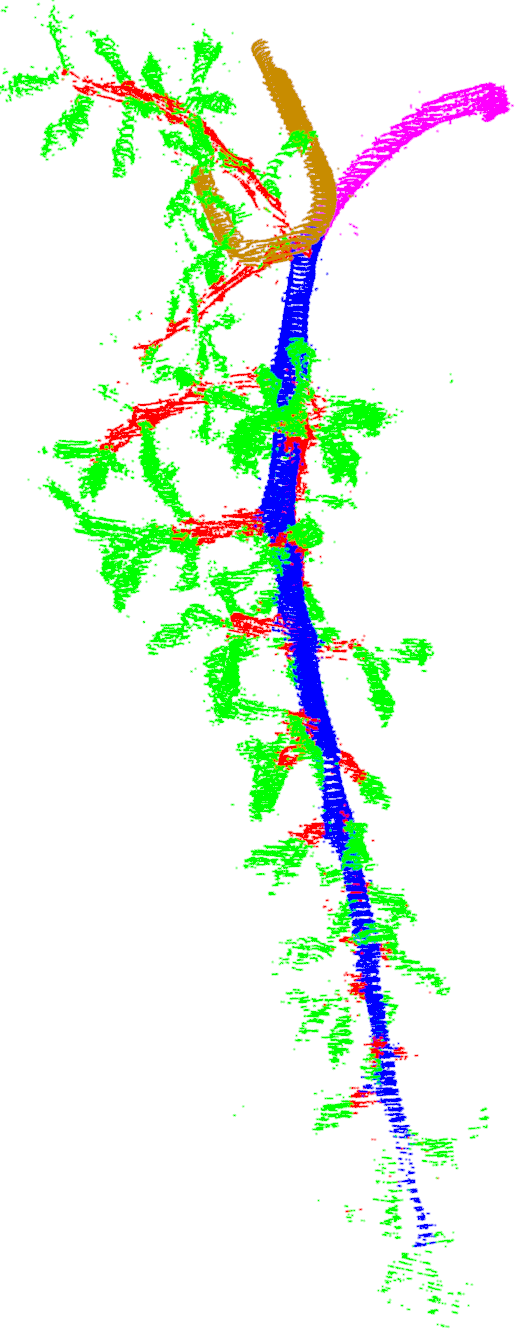}\label{fig_n6c}}
  \caption{The classes used in our more detailed semantic labeling of grape bunches with a distinction between twigs and sub-twigs. We take our RGB point clouds of grape bunches a) and apply the labeling scheme b) to create the labeled ground truth c).}
  \label{fig_nom6}
\end{figure}

\begin{figure}[!b]
\includegraphics[width=\textwidth]{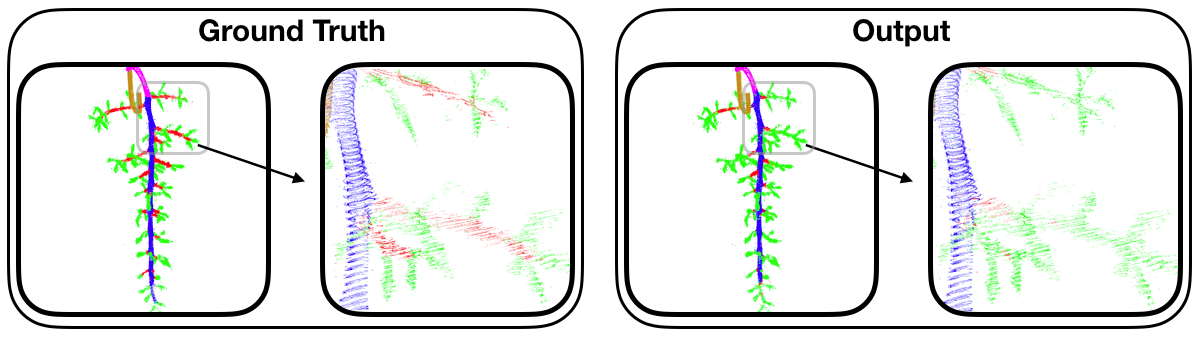}
\caption{Comparison of ground truth and output on a bunch of cultivar Dornfelder without berries. The labels shown are explained in Fig. \ref{fig_nom6}.}
\label{fig_compT}
\end{figure}

Figure \ref{fig_compB} compares the ground truth with the output of our model on a grape bunch with berries. At this task the main difficulty has been the distinction between berries and the hook. Due to the similar round shape other models tend to misclassify the hook as a berry. However, our model is capable of correctly identifying the berries as well as the hook apart from a few outliers. Even the barely visible stem structure has been labeled accordingly.

\begin{figure}[!h]
  \centering
  \subfloat[Normalized values]{\includegraphics[height=5cm]{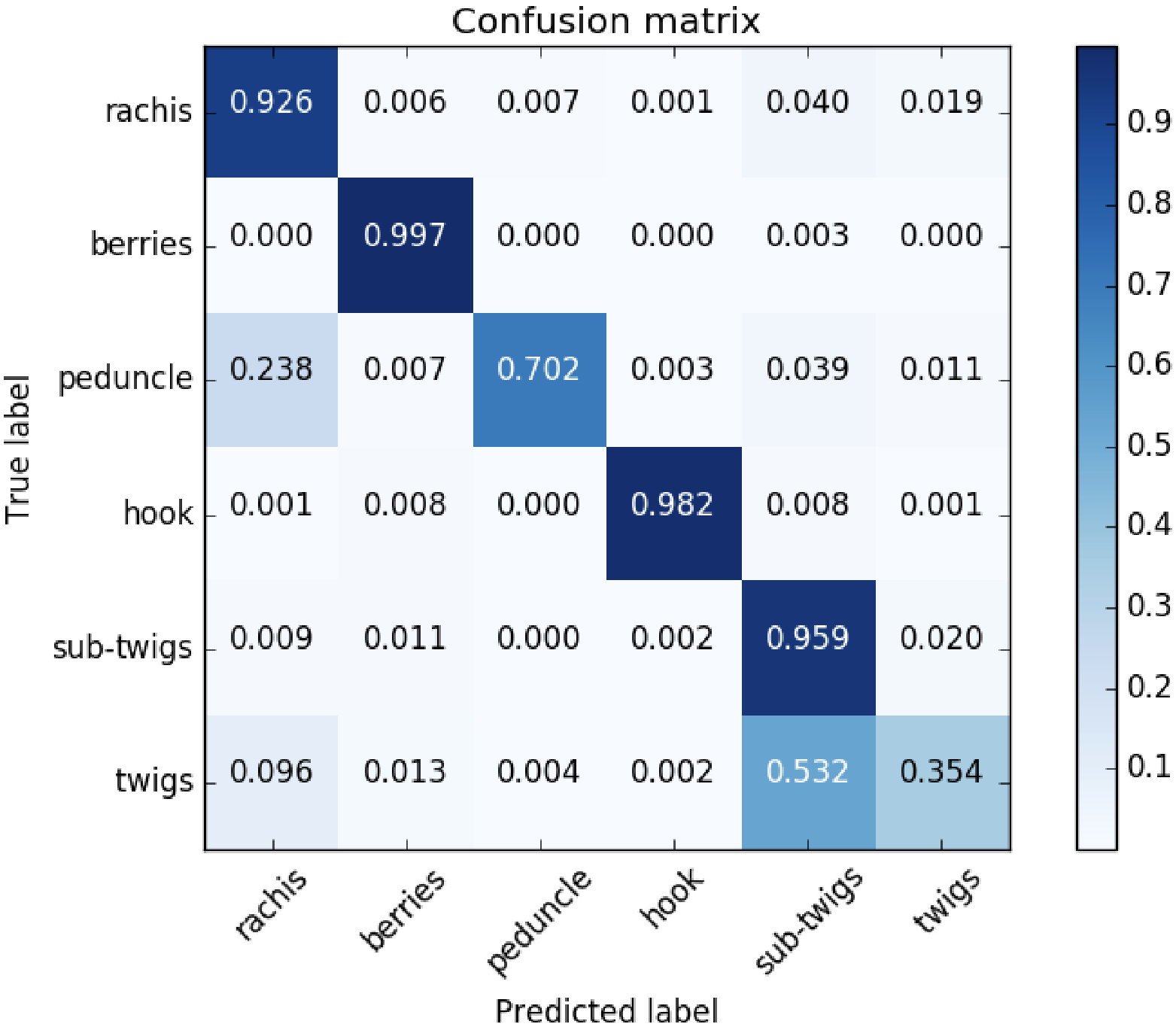}\label{fig_cm6norm}}
  \hfill
  \subfloat[Absolute values]{\includegraphics[height=5cm]{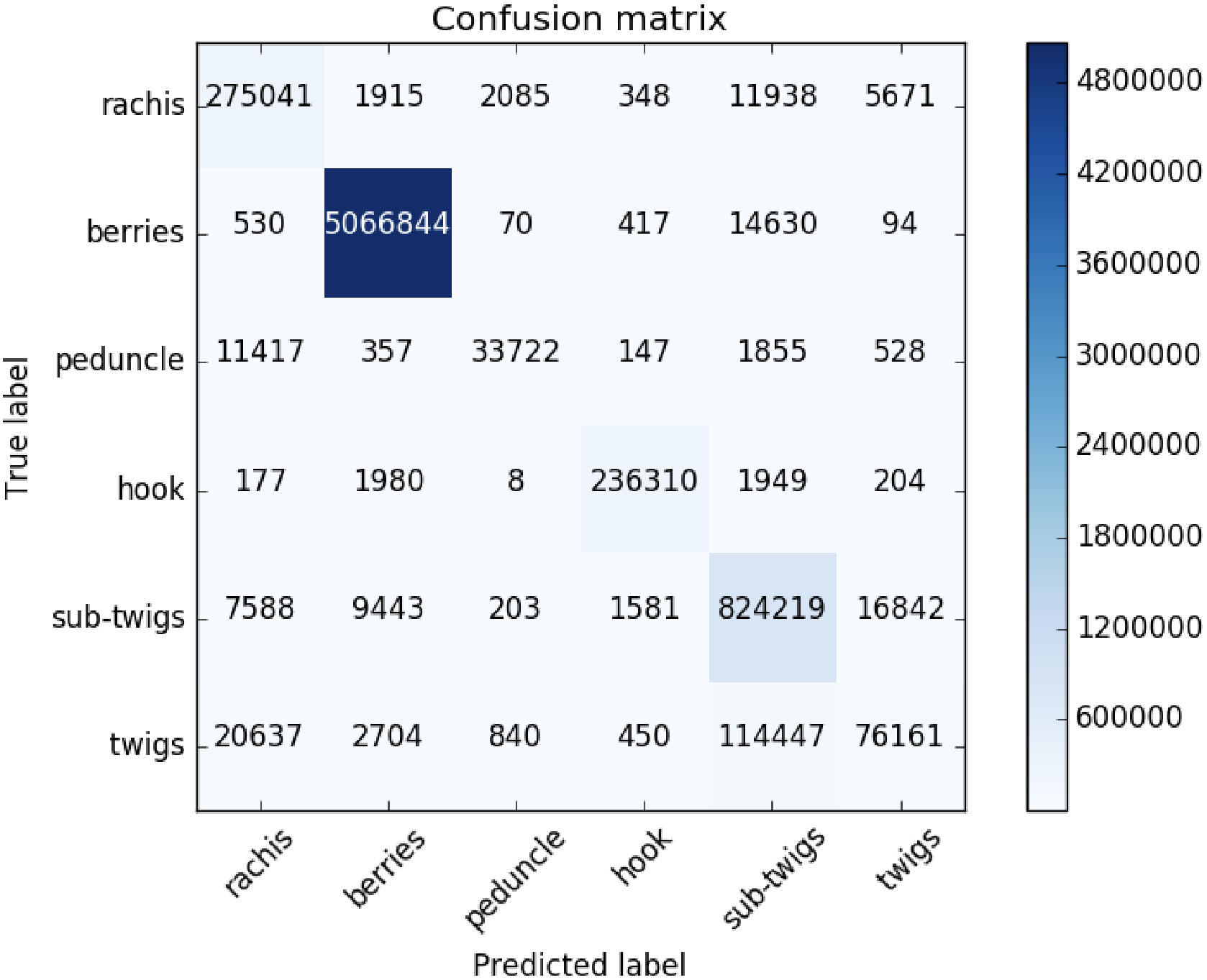}\label{fig_cm6abs}}
  \caption{Confusion matrix of the generated output for the whole set of test data using the detailed labeling scheme with six classes.}
  \label{fig_cm6both}
\end{figure}

For a more detailed phenotyping it would be useful to differentiate between twigs that are connected to the rachis and sub-twigs that are connected to twigs. In this case we would have six classes instead of five, namely: rachis, berries, peduncle, hook, twigs and sub-twigs. The resulting semantic labeling is shown in figure \ref{fig_nom6}. Our model can be easily modified to handle more classes by increasing the size of the last layers of our CNNs. Other than that the networks have been trained with the same parameters and settings as before.

The results are shown as a confusion matrix in figure \ref{fig_cm6both}. Our model is still able to detect and label rachis, berries and hook with high accuracies. However most of the combined twigs and sub-twigs are labeled as sub-twigs. Figure \ref{fig_compT} shows that in most cases only the connection of a twig and the rachis is labeled correctly. The part of the twig that is connected to multiple sub-twigs is not properly learned and in practice often times labeled as a sub-twig. This could be related to the distribution of our training data. The area of a twig that is connected to sub-twigs is often times hidden under the sub-twigs themselves. Our image generation for training data as described in section \ref{sec:Workflow} is sampling images at random with respect to the number of points. Instead it might be more efficient to use an advanced approach that is sampling based on class labels.

\section{Conclusions}

We proposed a multi-view based model for semantic labeling of 3D point clouds with applications in the automated plant phenotyping. So far our model has been trained and tested on high-resolution scans of grape bunches, data that has already been studied with regards to phenotyping by \citet{Stein15} and \citet{Mack17}.
Previous research relied on a combination of FPFH descriptors and an SVM to distinguish between the berries and stems of a grape bunch.

By using a multi-view approach and state-of-the-art CNN-based image segmentation our new model is able to learn efficient feature representations and their classification in order to identify berries as well as rachis, peduncle, twigs and the hook. The predictions on our test data have been very precise with an accuracy of 98.5\% and an AIoU of 87.7\%. The results on a more detailed phenotyping with a distinction between twigs and sub-twigs will be optimized with different image generation and sampling techniques in future work. 

We should further study how well our model generalizes on other grape cultivars and plants.
Currently our model relies on high-resolution point clouds taken under laboratory conditions with minimal amount of noise. Further research could focus on the applicability of our model on noisy data taken directly in the field. \citet{SnapNetR} have already proposed a new approach for multi-view semantic labeling in the context of robotics that could be combined with our results for plant phenotyping.

%
%
\bibliographystyle{plainnat}
\bibliography{multiview}
\end{document}